\documentclass[a4paper]{article}

\usepackage{INTERSPEECH2019}

\title{Performance Monitoring for End-to-End Speech Recognition}
\name{Ruizhi Li$^1$, Gregory Sell$^{1,2}$, Hynek Hermansky$^{1,2}$ \thanks{This material is based upon work supported by the National Science Foundation under Grant No. 1704170 and No.  1743616.}}
\address{
  $^1$Center for Language and Speech Processing, The Johns Hopkins University, USA\\
  $^2$Human Language Technology Center of Excellence, The Johns Hopkins University, USA}
\email{\{ruizhili, gsell2, hynek\}@jhu.edu}

\begin{document}
 
\maketitle

\begin{abstract}
Measuring performance of an automatic speech recognition (ASR) system without ground-truth could be beneficial in many scenarios, especially with data from unseen domains, where performance can be highly inconsistent. 
In conventional ASR systems, several performance monitoring (PM) techniques have been well-developed to monitor performance by looking at tri-phone posteriors or pre-softmax activations from neural network acoustic modeling.  However, strategies for monitoring more recently developed end-to-end ASR systems have not yet been explored, and so that is the focus of this paper.
We adapt previous PM measures (Entropy, M-measure and Auto-encoder) and apply our proposed RNN predictor in the end-to-end setting. 
These measures utilize the decoder output layer and attention probability vectors, and their predictive power is measured with simple linear models. 
Our findings suggest that decoder-level features are more feasible and informative than attention-level probabilities for PM measures, and that M-measure on the decoder posteriors achieves the best overall predictive performance with an average prediction error 8.8\%.  Entropy measures and RNN-based prediction also show competitive predictability, especially for unseen conditions.

\end{abstract}

\section{Introduction}
In recent years, significant improvement for conventional automatic speech recognition (ASR) has been achieved via advancements with Deep Neural Networks (DNNs).
The main paradigm for an ASR system is the so-called hybrid approach, which involves training a DNN to predict context dependent phoneme states (or senones) from the acoustic features.
However, if test data comes from a very different domain than DNN training data, the recognizer is likely to fail without any warning. Predicting these failures is the goal of the work that follows.
Humans are often aware of the uncertainty of decisions they are making \cite{humanPM}.
% able to evaluate the reliability of a transcription without hearing the original speech.
Performance monitoring (PM) techniques aim for the same goal - to determine the quality of a system's output - based only on the behavior of the system and without any knowledge of the underlying truth. An effective PM measure could be useful in a number of applications \cite{ganapathy2013unsupervised,okawa1998multi,mallidi2016framework,DBLP:journals/speech/MartinezGVHOM19,wang2018stream,wang2018stream1, meyer2016performance, hermansky2013multistream}, such as multi-stream selection scenario \cite{mallidi2016framework, wang2018stream, hermansky2013multistream} or semi-supervised training \cite{ganapathy2013unsupervised}.

Unlike conventional ASR, end-to-end speech recognition approaches are designed to directly output word or character sequences from the input audio signal. 
This model subsumes several disjoint components in the hybrid ASR model (acoustic model, pronunciation model, language model) into a single neural network.  
As a result, all the components of an end-to-end model can be trained jointly to optimize a single objective.  
Two dominant end-to-end architectures for ASR are Connectionist Temporal Classification (CTC) \cite{graves2006connectionist,graves2014towards,miao2015eesen} and attention-based encoder-decoder models \cite{chan2015listen,chorowski2015attention} .   
A joint CTC/Attention framework was proposed in \cite{kim2016joint_icassp2017,hori2017advances,watanabe2017hybrid}.  to take advantage of both architectures within a multi-task scheme.
The joint model was shown to provide state-of-the-art end-to-end results for several benchmark datasets \cite{watanabe2017hybrid,li2018multi}. 

This paper aims to explore the PM techniques applicable for an end-to-end framework. 
In the hybrid approach, tri-phone posterior distributions and their corresponding pre-softmax activations are typically as PM features. 
Averaged entropy over temporal frames was proposed as a confidence measure in stream-selection \cite{okawa1998multi,misra2003new}. 
Mean temporal distance on posteriors estimates the performance by capturing the divergence of any two frames over several time spans \cite{variani2013multi, hermansky2013mean}.
Reconstruction error of an auto-encoder trained on pre-softmax features was also used as the selection criterion in a multi-stream system \cite{mallidi2015uncertainty, mallidi2015autoencoder}. 

In the end-to-end setting, there are two levels of probability distributions: attention weights and decoder posteriors. 
Instead of temporal posteriors in the conventional case, each probability distribution corresponds to a character-level prediction.
Therefore, we must adapt the techniques used for hybrid systems to the joint CTC/attention model. 
Moreover, inspired  by the success of discriminatively-trained DNNs, we propose using a Recurrent Neural Network (RNN) regression model trained to directly predict performance. 
Our analyses demonstrate strong correlations between PM measures and true performance, indicating that end-to-end ASR systems are indeed amenable to effective monitoring.

This paper is organized as follows: Section 2 explains the joint CTC/Attention model. The description of data configuration is in Section 3. 
Experiments with results and analyses are presented in Section 4. Finally, we conclude in Section 5. 

\section{End-to-End Architecture}

The joint CTC/Attention model is designed to directly map $T$-length acoustic features $X=\{\textbf{x}_{t}\in \mathbb{R}^{D}|t = 1,2,...,T\}$ in $D$ dimensional space to an $L$-length letter sequence $C=\{c_{l}\in \mathcal{U}|l = 1,2,...,L\}$ where $\mathcal{U}$ is a set of distinct letters. 
The attention-based structure solves the ASR problem as a sequence mapping by using an encoder-decoder architecture. 
Joint training with CTC is added to help enforce temporally monotonic behavior in the attention alignments. 

\begin{figure}[tb]
  \centering
  \includegraphics[width=0.6\linewidth]{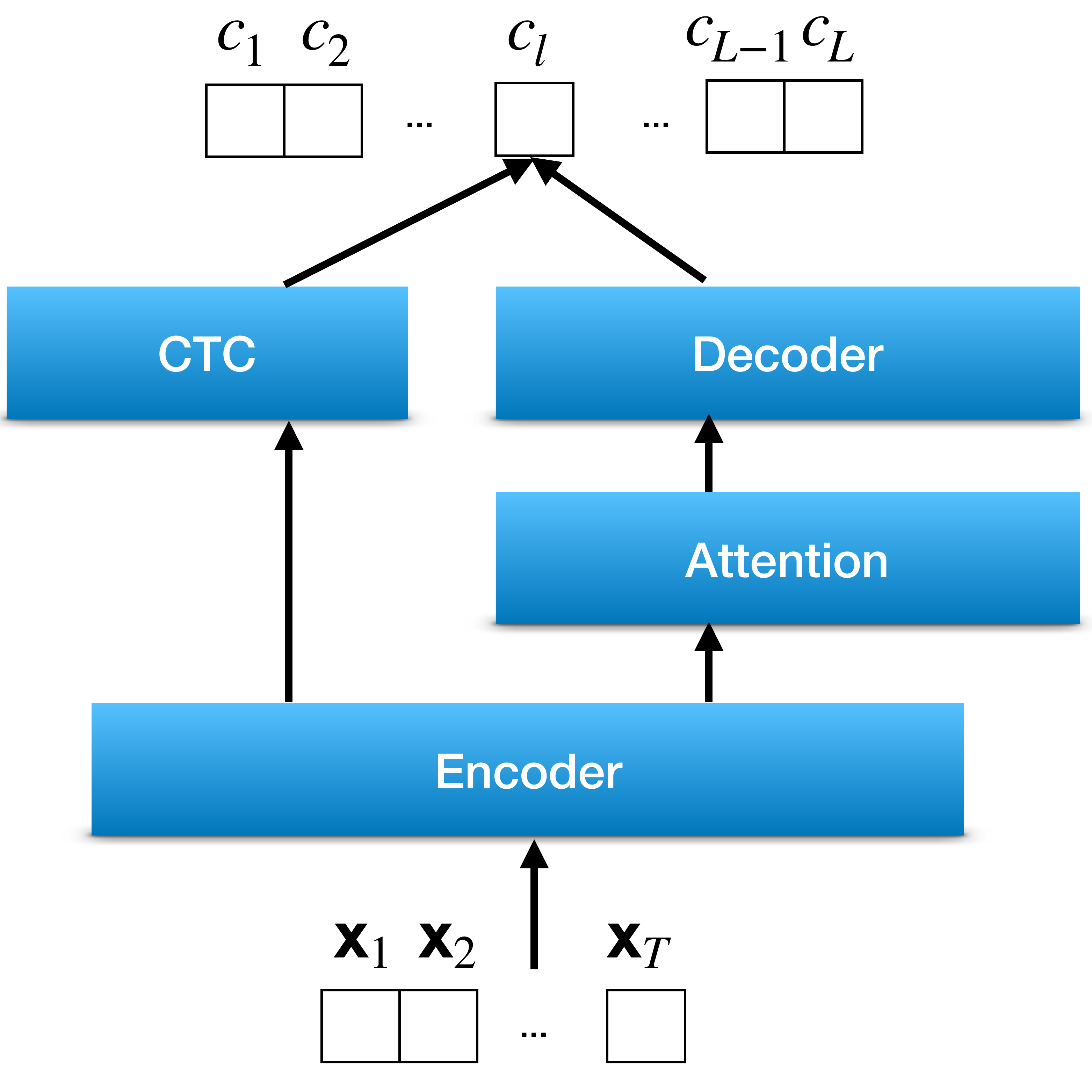}
  \caption{Joint CTC/Attention End-to-End Architecture}
  \label{fig:e2e}
\end{figure}

The overall end-to-end architecture is shown in  Fig. \ref{fig:e2e}. 
The encoder is shared by both the attention and CTC networks.
Bidirectional Long Short-Term Memory (BLSTM) layers are utilized to model the temporal dependencies of the input sequence. 
The frame-wise hidden vector $\textbf{h}_{t}$ at frame $t$ is derived by encoding the full input sequence $X$:
\begin{equation}
\textbf{h}_{t}=\textrm{Encoder}(X) 
\end{equation}
In the attention-based network, the letter-wise context vector $\textbf{r}_{l}$ is formed as a weighted summation of frame-wise hidden vectors $\textbf{h}_{t}$ using a content-based attention mechanism:
\begin{equation}
\textbf{r}_{l}={\sum}_{t=1}^{T}a_{lt}\textbf{h}_{t},\ {a}_{lt}=\textrm{ContentAttention}(\textbf{q}_{l-1}, \textbf{h}_t) 
\end{equation}
where $\textbf{q}_{l-1}$ is the previous decoder state, and ${a}_{lt}$ is the attention weight, which can be considered a soft-alignment of $\textbf{h}_{t}$ for $c_{l}$. 
An LSTM-based decoder outputs the character-level conditional probability distribution $p(c_{l}|c_{1},...,c_{l-1}, X)$ for multi-task training and joint decoding.

During inference, the joint CTC/Attention model performs a label-synchronous beam search, which jointly predicts the next character by considering a CTC network and an attention decoder.
No external language model is involved in this work. 
The most probable letter sequence $\hat C$ given the speech input $X$ is computed as:
\begin{align}
\label{equ:max}
\hat{C}=\arg\max_{C\in \mathcal{U}^{*}} &\{\lambda \log p_{ctc}(C|X)+(1-\lambda)\log p_{att}(C|X)\}
 \end{align}
where we can directly estimate the attention posterior distribution $p_{att}(C|X)$ using the chain rule:
\begin{equation}
p_{att}(C|X)=\coprod_{l=1}^{L}p(c_{l}|c_{1},...,c_{l-1}, X).
\end{equation}

In this work, we apply PM techniques on three kinds of features to predict the CERs: attention distributions $\{a_{lt}\}$, decoder posterior distributions $p(c_{l}|c_{1},...,c_{l-1}, X)$, and pre-softmax activations of $p(c_{l}|c_{1},...,c_{l-1}, X)$.

\section{Data}
% train 
We conduct all our experiments based on the Wall Street Journal (WSJ) corpus \cite{wsj1} and its variants with additional noises or reverberation conditions.   
Table \ref{tab:datainfo} summarizes various databases that are used in our experiments. 
For end-to-end ASR, the clean WSJ SI-284 corpus and Aurora4  multi-condition training data are used in multi-style training.
The Aurora4 set \cite{pearce2002aurora} contains simulated recordings of WSJ utterances in 14 different acoustic conditions, varying noise types and channel conditions. 
WSJ dev93 and Aurora4 dev set serve as the validation set for ASR training. 
We also use the same data configuration to train the auto-encoder. 
For the RNN predictor, in order to see data with a reasonable balance across different CERs, we use clean WSJ SI-84 together with its two artificially noise-corrupted versions, Aurora4 and CHiME4-Sim. CHiME4-Sim \cite{vincent20164th} consists of single-channel WSJ data with four additive noises. 

\begin{figure}[!tbp]
\begin{minipage}[b]{.32\linewidth}
  \centering
  \centerline{\includegraphics[width=2.8cm]{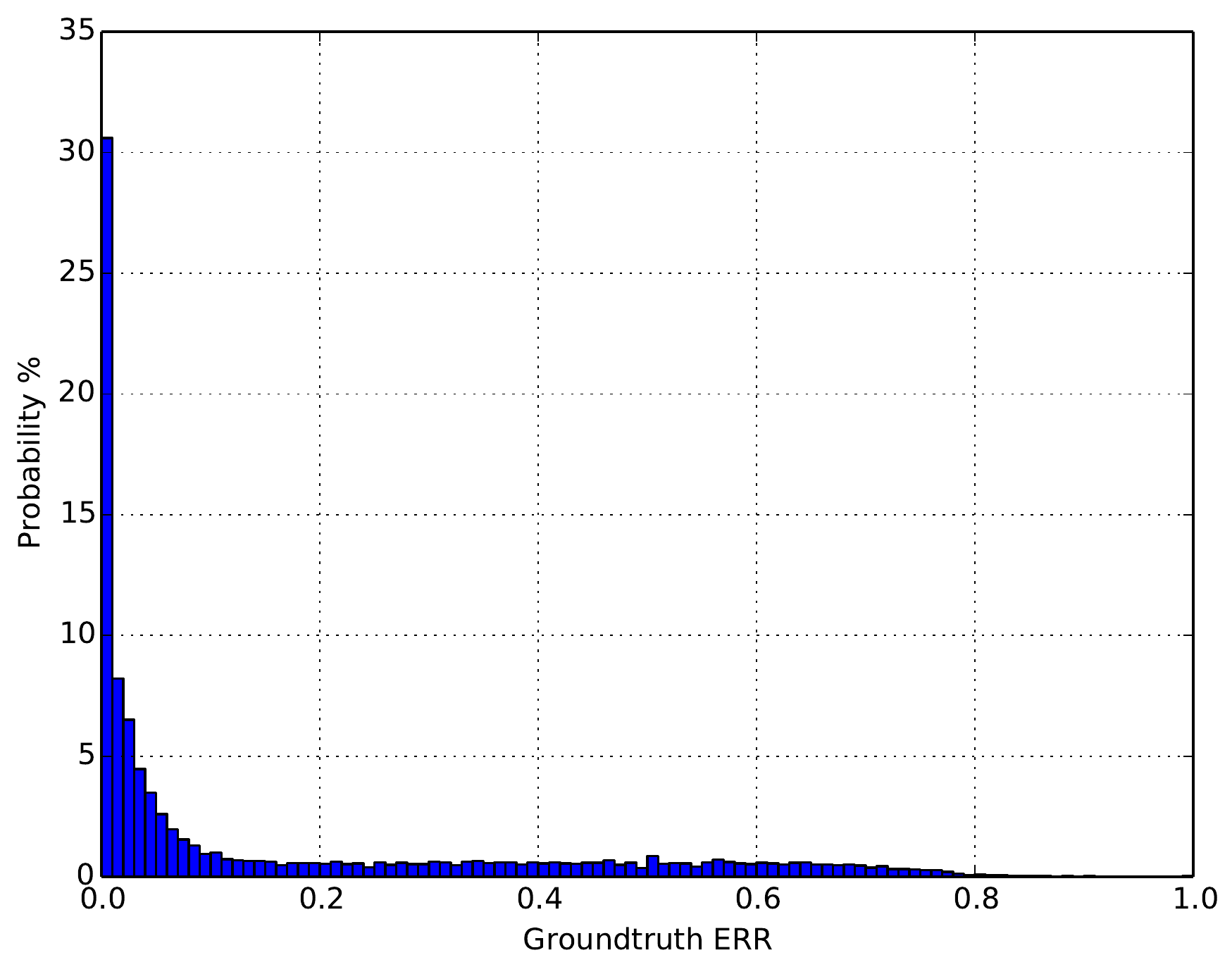}}
  \centerline{(a) Train}\medskip
\end{minipage}
\hfill
\begin{minipage}[b]{.32\linewidth}
  \centering
  \centerline{\includegraphics[width=2.8cm]{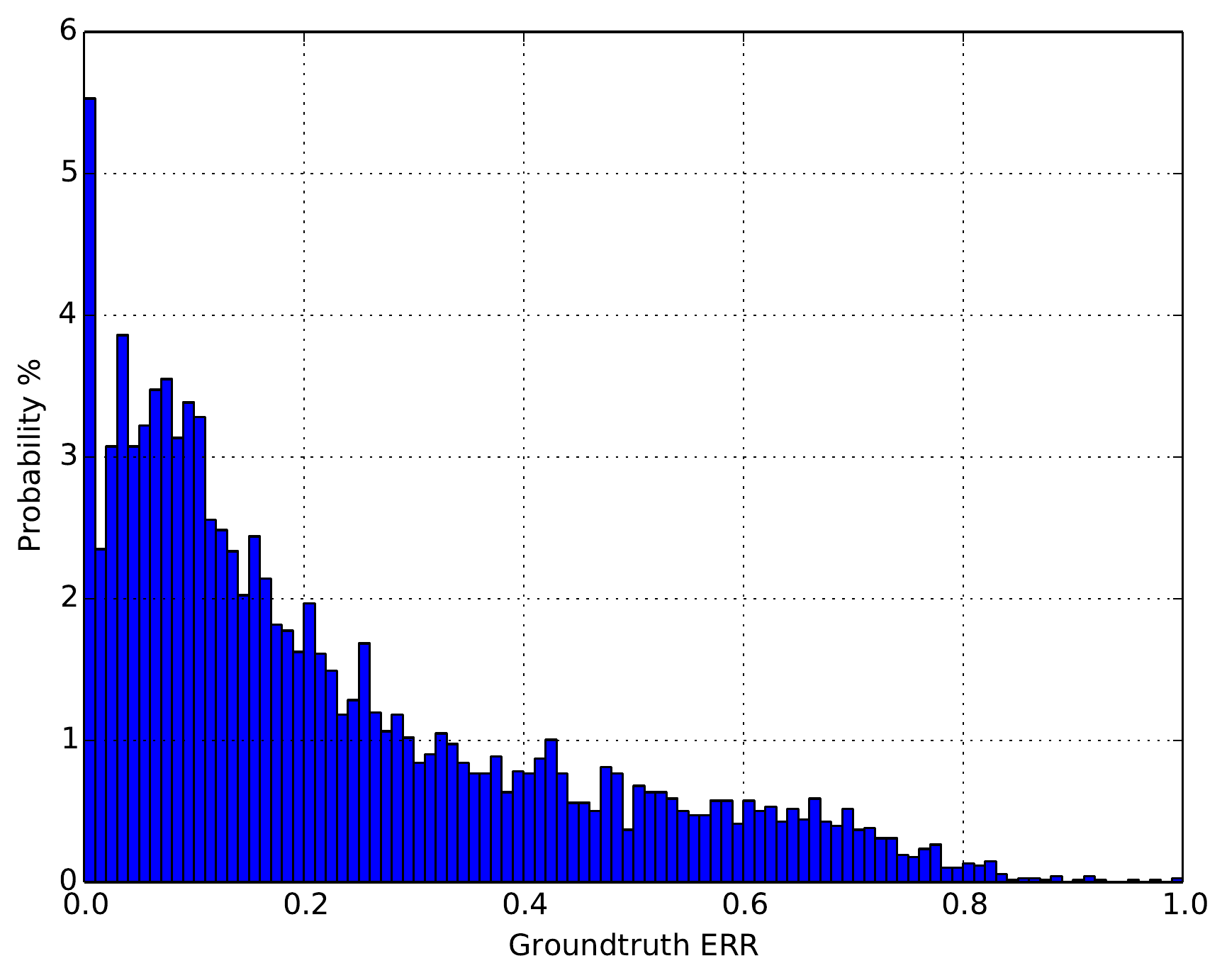}}
  \centerline{(b) Dev}\medskip
\end{minipage}
\hfill
\begin{minipage}[b]{0.32\linewidth}
  \centering
  \centerline{\includegraphics[width=2.8cm]{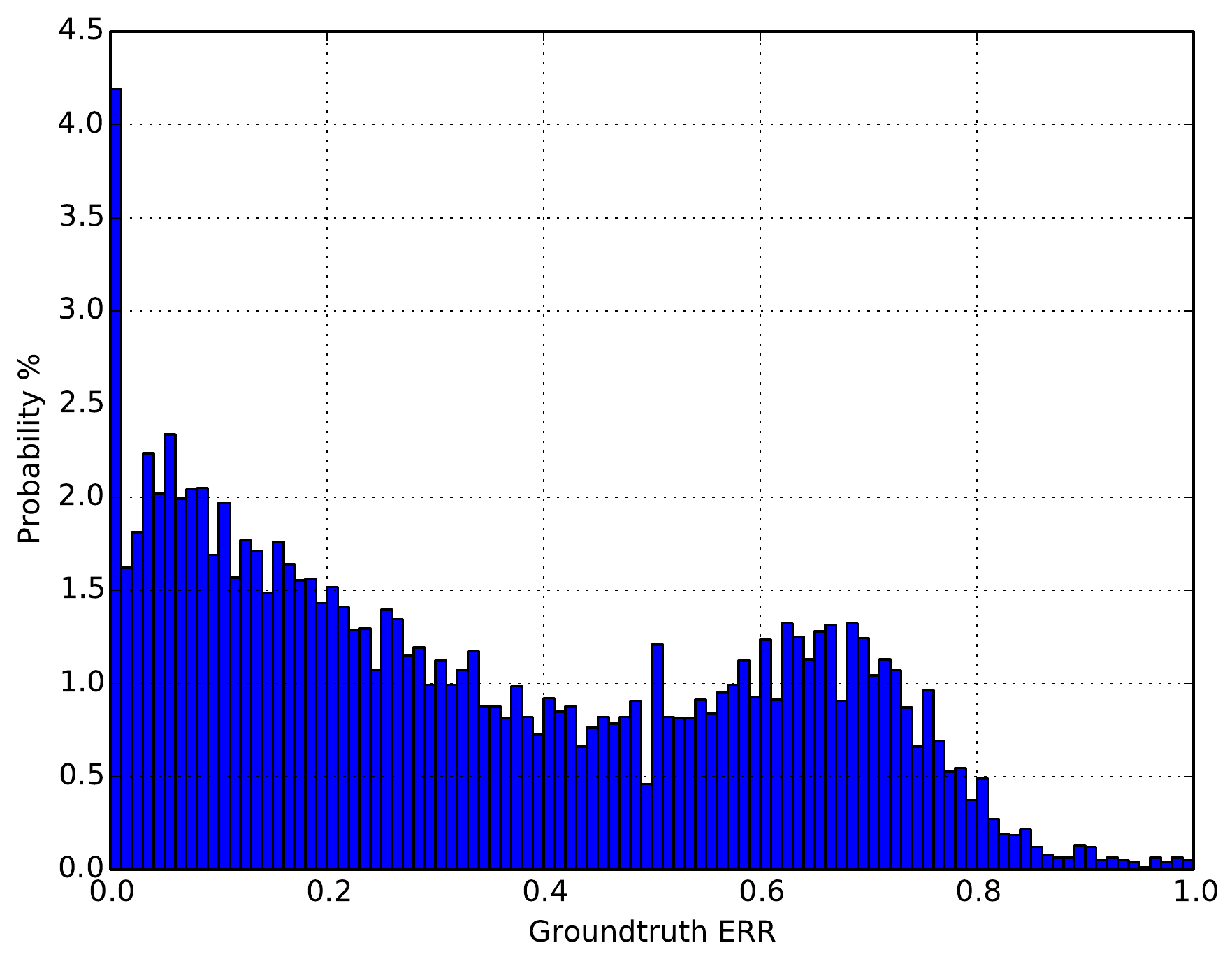}}
  \centerline{(c) Test}\medskip
\end{minipage}
\caption{Histogram of CERs in various datasets}
\label{fig:hist}
\end{figure}

% dev 
To evaluate the predictability of each PM measure, a linear regression model is applied to map between PM scores and truth performance. The regression model is computed according to development sets from WSJ, Aurora4 and CHiME4-Sim. We refer the data to train the ASR and linear regression model as \textit{Train} and \textit{Dev}, respectively. Fig. \ref{fig:hist}(a) and Fig. \ref{fig:hist}(b) show histograms of utterance CERs for both sets. Performance of \textit{Dev} is more widely spread out than \textit{Train}. 
% test 
We test the effectiveness of PM measures with data from two different domains. \textit{Seen} test domains include evaluation sets from WSJ, Aurora4 and CHiME4-Sim which are drawn from the same domains as ASR and PM training; \textit{Unseen} test domains consist of evaluation sets from CHiME4-Real \cite{vincent20164th} (real noisy recordings), Reverb-Sim \cite{6701894} (simulated reverberation), Dirha-Sim \cite{ravanelli2016realistic} (simulated reverberation) and Dirha-Real ]\cite{ravanelli2016realistic} (real reverberated recordings). All the test data together are referenced as \textit{Test}, with utterance CERs shown in Fig. \ref{fig:hist}(c). 

\begin{table}[tb]
  \caption{Datasets for experiments with CER (\%) of Test set}
  \label{tab:datainfo}
  \centering
  \resizebox{0.42\textwidth}{!}{\begin{tabular}{lll}
  \toprule
  \toprule
  Task & Dataset\\
  \midrule
  {\it Train} \\
  ASR & \multicolumn{2}{l}{WSJ(SI-284), Aurora4}\\
  AE & \multicolumn{2}{l}{WSJ(SI-284), Aurora4} \\
  RNN & \multicolumn{2}{l}{WSJ(SI-84), Aurora4, Chime4-Sim} \\
  \midrule
  {\it Dev} \\
  Linear Regr & \multicolumn{2}{l}{WSJ, Aurora4, Chime4-Sim}\\
  
  \midrule
  {\it Test} \\
  All Tasks& WSJ (5.7\%) & Aurora4 (14.5\%)  \\
  & Chime4-Sim (52.2\%) & Dirha-Sim (68.2\%)  \\
  & Chime4-Real (59.0\%) & Dirha-Real (70.7\%) \\
  & Reverb (41.6\%) &  \\
  \bottomrule 
  \bottomrule
  \end{tabular}}
\end{table}

\section{Experiments}

\subsection{Experiment Setup}

\begin{figure*}
\begin{minipage}[b]{0.33\linewidth}
  \centering
  \centerline{\includegraphics[width=5.0cm]{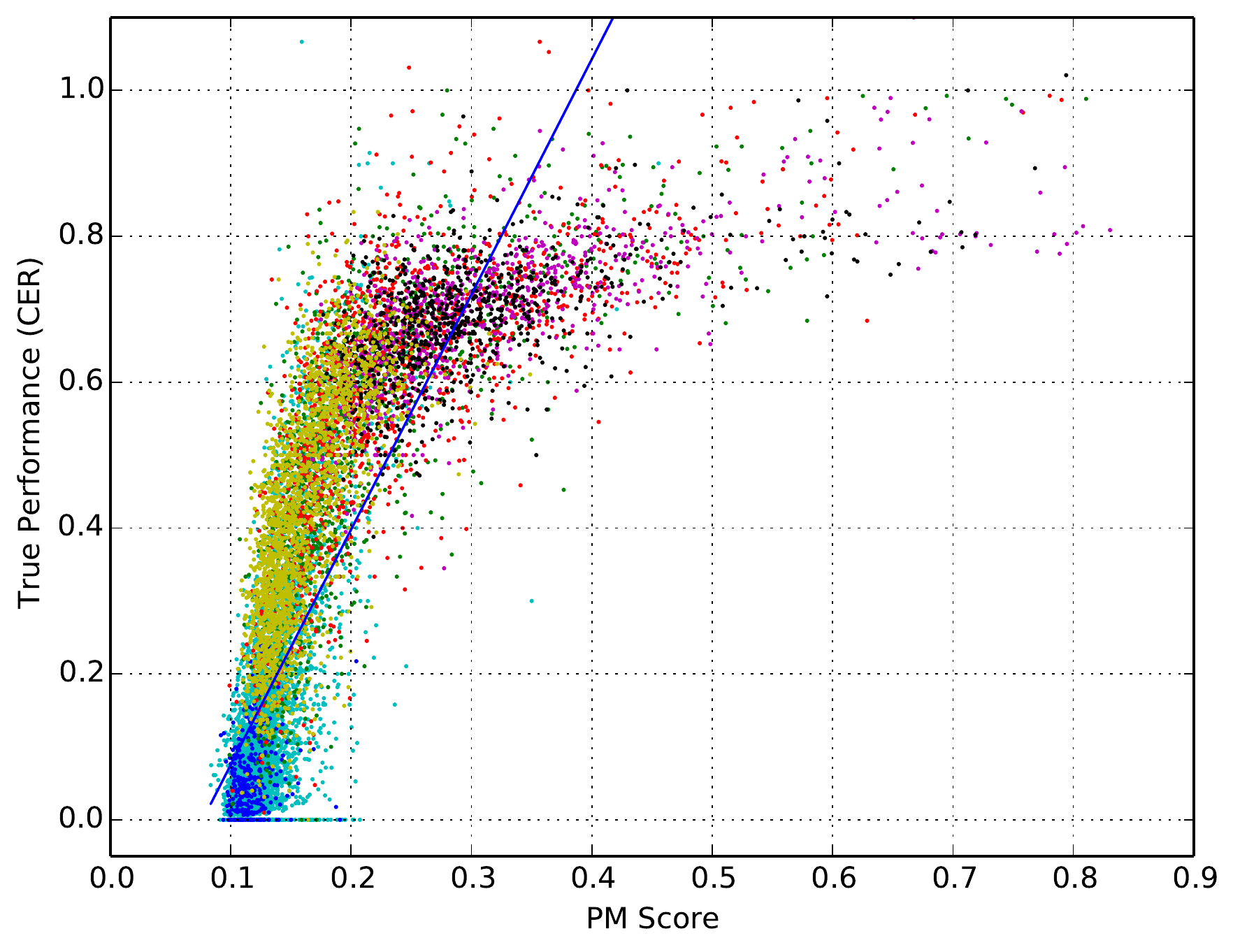}}
  \centerline{(a) Entropy (Attention Posteriors)}\medskip
\end{minipage}
\hfill
\begin{minipage}[b]{0.33\linewidth}
  \centering
  \centerline{\includegraphics[width=5.0cm]{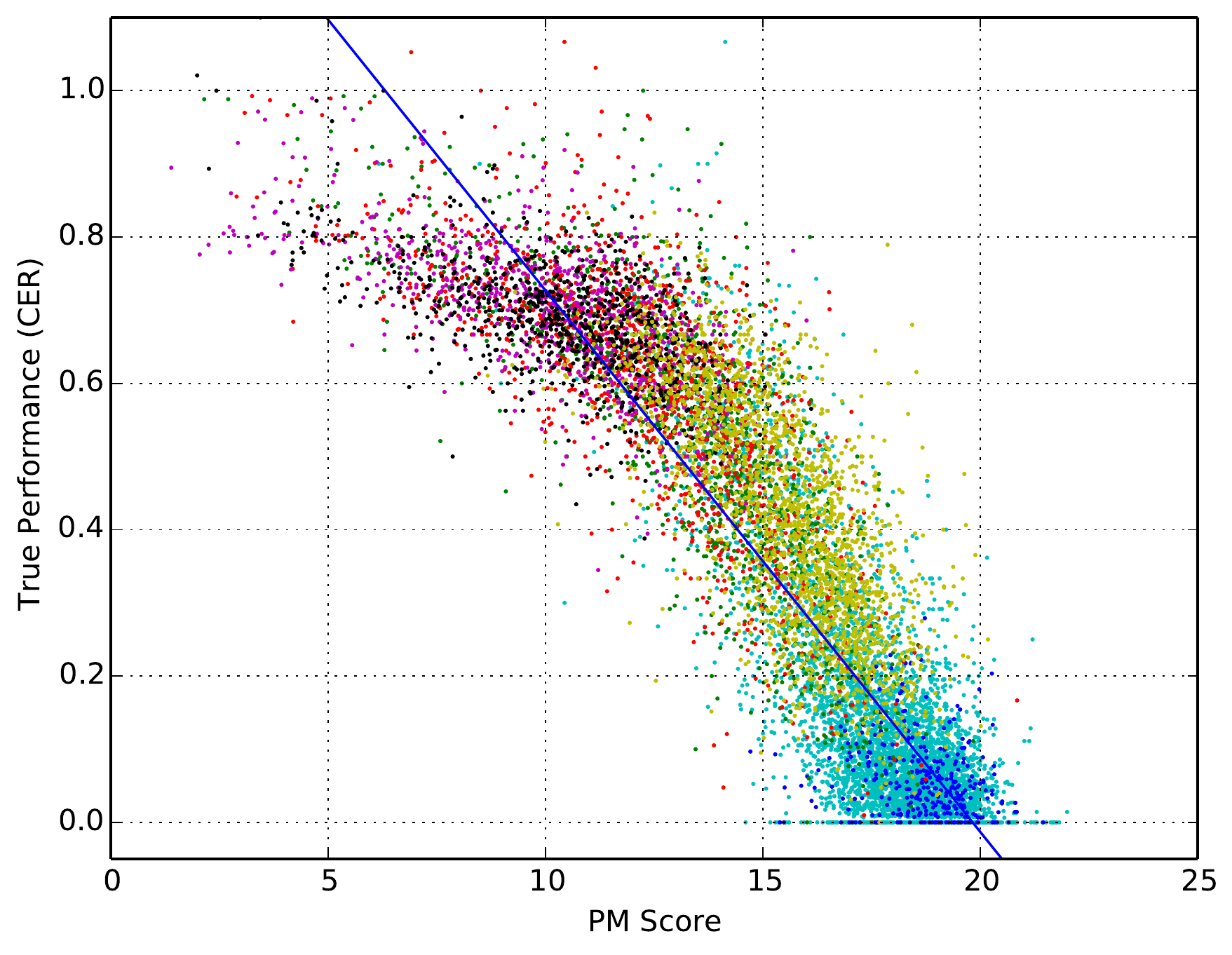}}
  \centerline{(c) MCD (Attention Posteriors)}\medskip
\end{minipage}
\hfill
\begin{minipage}[b]{0.33\linewidth}
  \centering
  \centerline{\includegraphics[width=5.0cm]{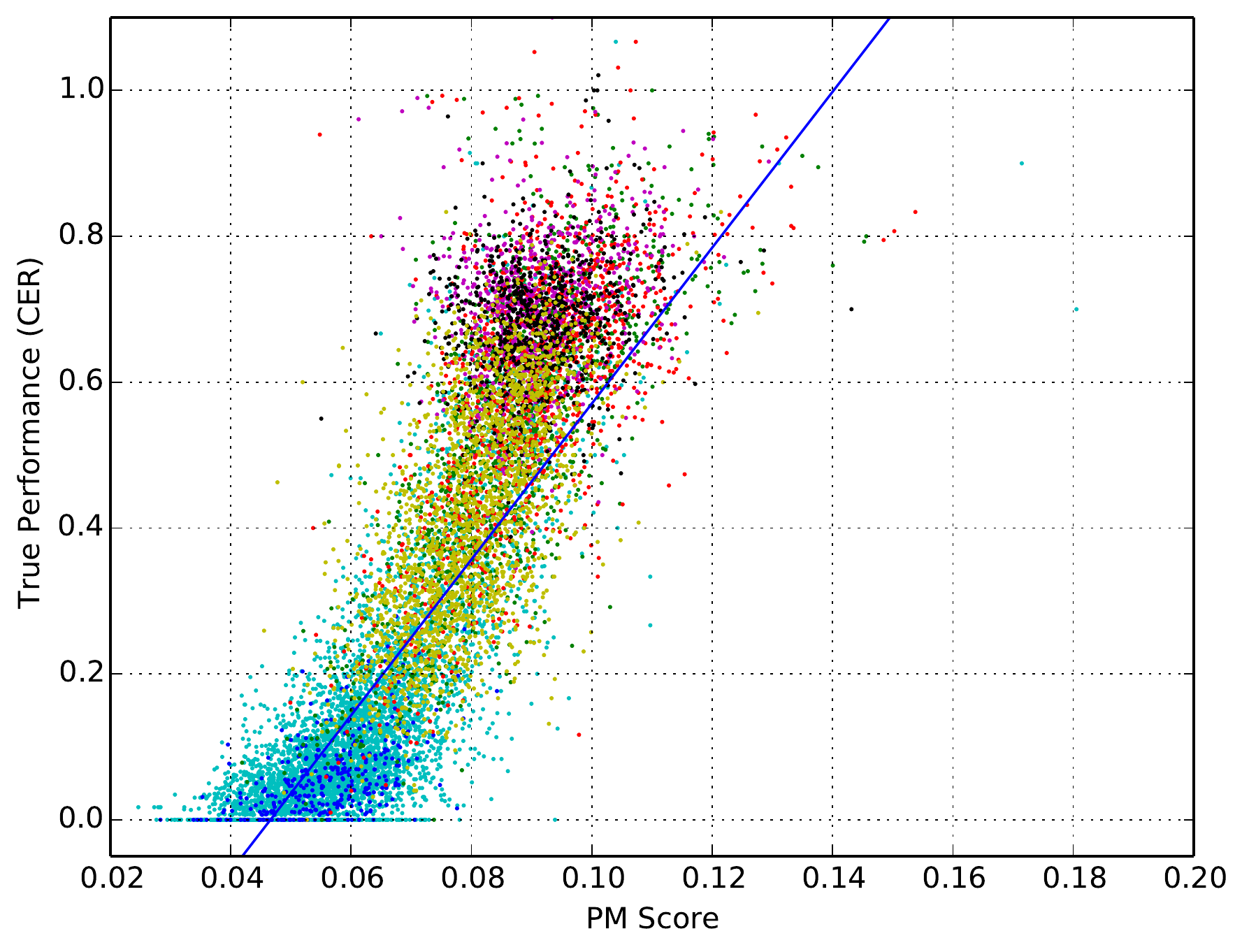}}
  \centerline{(e) Auto-encoder (Decoder Pre-softmax)}\medskip
\end{minipage}

\begin{minipage}[b]{0.33\linewidth}
  \centering
  \centerline{\includegraphics[width=5.0cm]{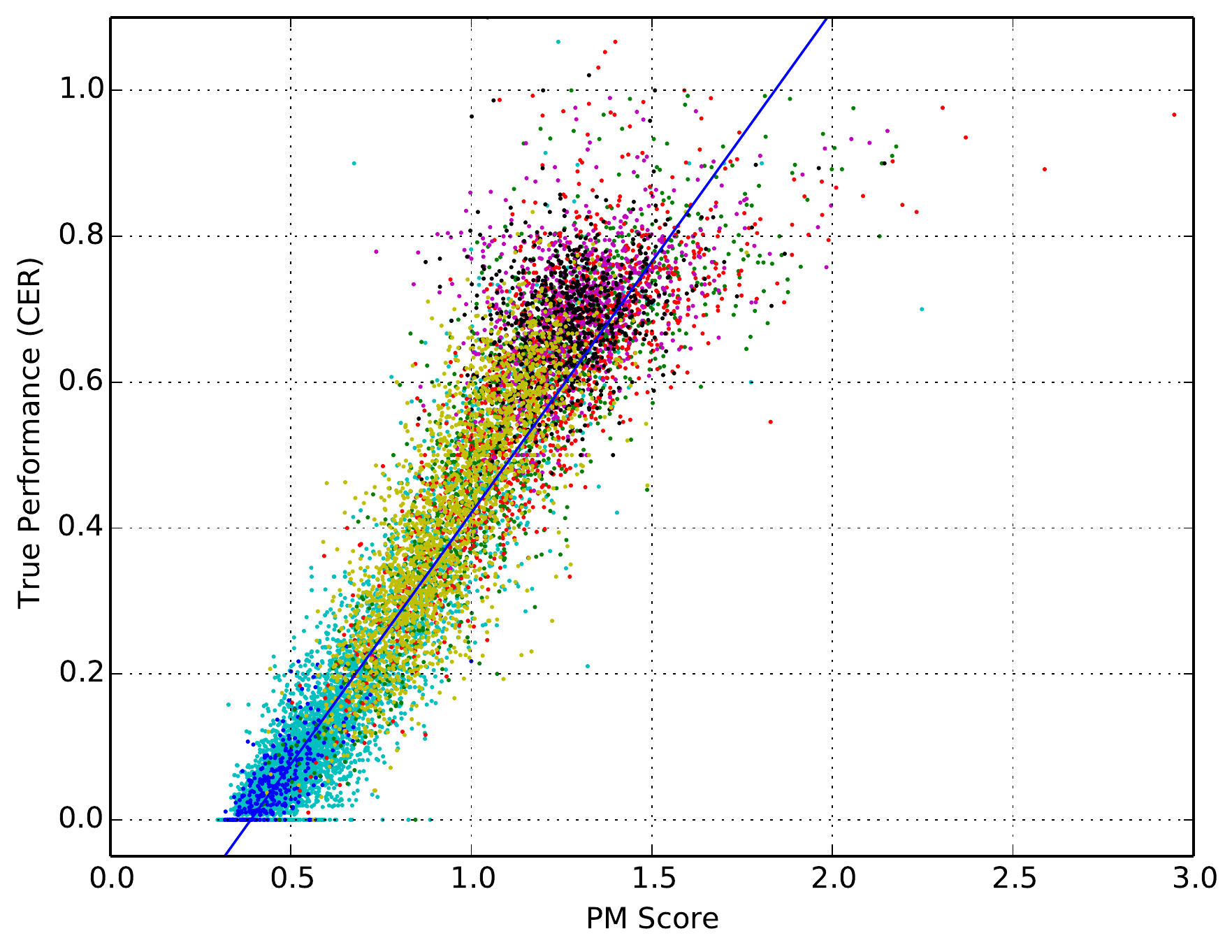}}
  \centerline{(b) Entropy (Decoder Posteriors)}\medskip
\end{minipage}
\hfill
\begin{minipage}[b]{0.33\linewidth}
  \centering
  \centerline{\includegraphics[width=5.0cm]{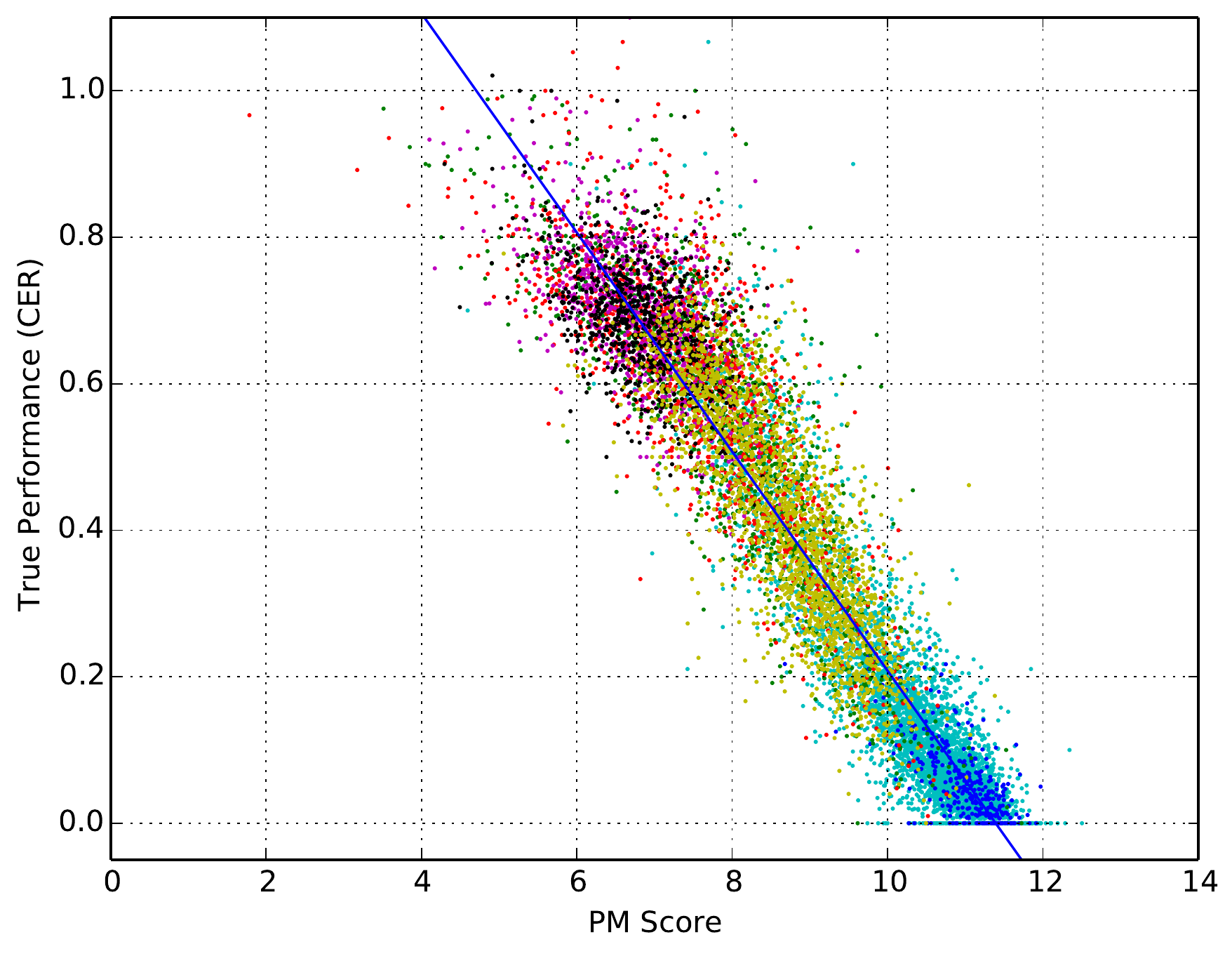}}
  \centerline{(d) MCD (Decoder Posteriors)}\medskip
\end{minipage}
\hfill
\begin{minipage}[b]{0.33\linewidth}
  \centering
  \centerline{\includegraphics[width=5.0cm]{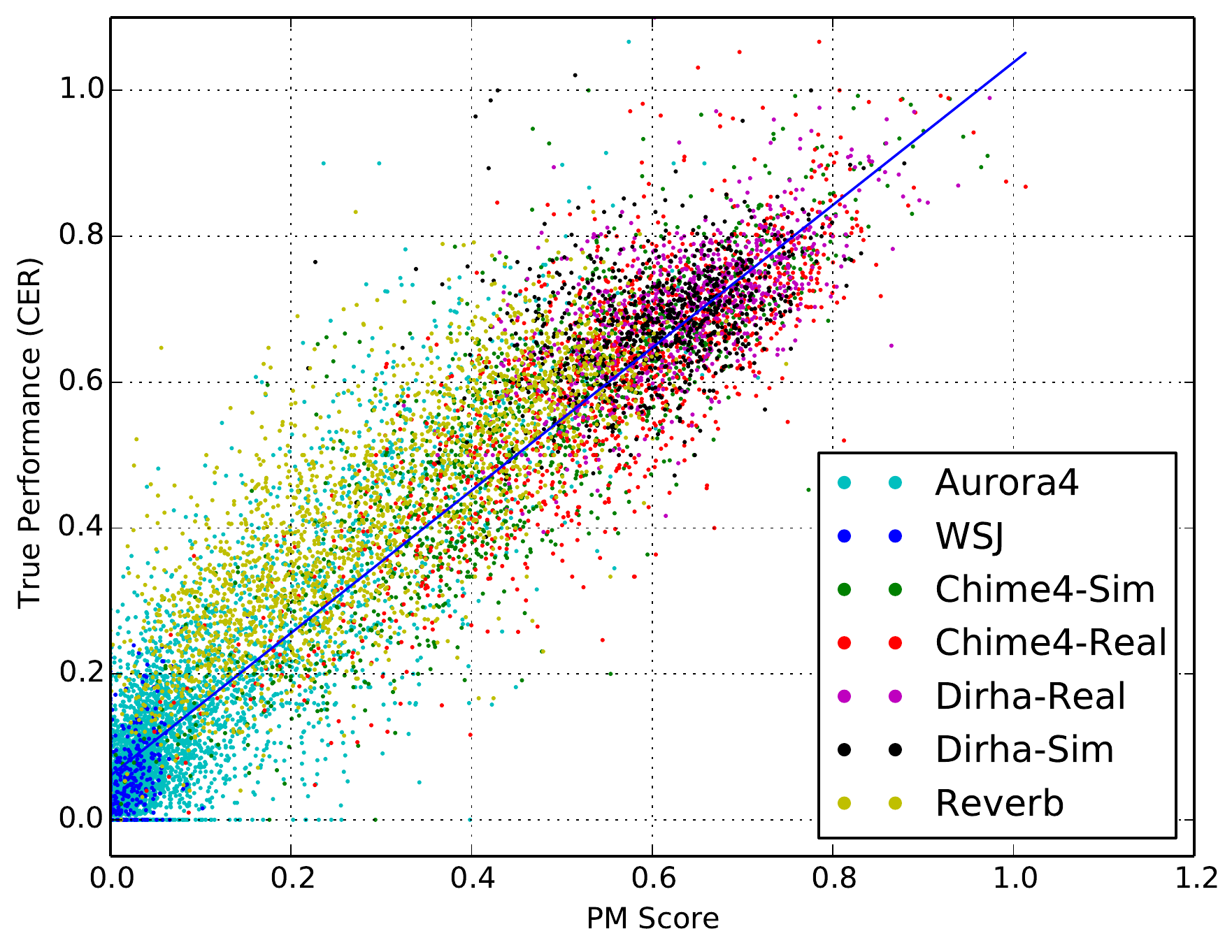}}
  \centerline{(f) RNN Predictor (Decoder Pre-softmax)}\medskip
\end{minipage}

\caption{{Performance Monitoring Score versus Truth Performance (CER)}}
\label{fig:mse}
\end{figure*}

In the end-to-end model, the encoder contains four BLSTM layers, each with 320 cells in both directions, followed by a 320-unit linear projection layer. 
A content-based attention mechanism with 320 attention units follows. 
The decoder is a one-layer unidirectional LSTM with 300 cells. 
We use 52 distinct labels at the decoder softmax layer, including 26 English letters and additional special tokens (i.e., punctuations and sos/eos). 

The model is implemented using the Pytorch backend on ESPnet \cite{watanabe2018espnet}, an end-to-end speech processing toolkit. 
The model is optimized using the AdaDelta algorithm with a mini-batch size of 15.
We also apply a unigram label smoothing technique to avoid over-confident predictions.
The beam width is set to 30 for all results. 
For jointly training with CTC and attention objectives, $\lambda=0.2$ is used for training, and $\lambda=0.3$ for decoding.
All results are reported as CER.
In all experiments, 80-dimensional mel-scale filter-bank coefficients with additional 3-dimensional pitch features served as the input features. Attention distributions, decoder posteriors, and pre-softmax features are extracted during joint decoding. Decoding results of each \textit{Test} set are shown in Table. \ref{tab:datainfo}.

\subsection{Entropy}

In the hybrid ASR framework, it was observed \cite{okawa1998multi,misra2003new} that the discriminative power of a clean phoneme classifier decreases for input speech with reduced signal-to-noise ratios.  
As the phoneme posteriors tend to be more uniformly distributed, entropy was proposed as a measure of uncertainty.
In end-to-end ASR, which has no phoneme distributions, we investigate if entropy on either the attention probabilities or the decoder posteriors could be a reasonable indicator of model performance. 

Entropy is first computed on the character-level distribution $p(c_l|c_{1},...,c_{l-1}, X)$. 
\begin{equation}
\label{f:ent}
Entropy(\mathbf{p}) = -\sum_{k=0}^{K} p^{(k)}\log{p^{(k)}} 
\end{equation}
where $\textbf{p}_l$ is either the attention probabilities $\textbf{a}_{l}$ or decoder posteriors.
The  utterance-level score is obtained by averaging entropy scores over all predictions. 
\begin{equation}
\label{f:escore}
E_{score} = \frac{1}{L}\sum_{l=0}^{L} Entropy(\textbf{p}_{l}) 
\end{equation}
Note that the dimension of the attention distribution is equal to the number of time frames $T$ at encoder output, which varies per utterance.  So, we normalize this entropy by its upper bound $\log (T)$ for a consistent range $[0, 1]$. Fig. \ref{fig:mse}(a)(b) show scatter plots of entropy scores versus truth CERs on \textit{Test}, where each point in the plot represents one utterance.
A linear model $CER = a*PM+b$ learned to minimize the mean squared error over \textit{Dev} is also shown. 
Entropy scores on the decoder output clearly demonstrate a linear relationship with true performance, while linear correlation for attention-level distributions holds only for error rates less than 0.5, resulting in larger prediction error overall. 

\subsection{M-Measure: Mean Character Distance}

The Mean Temporal Distance (MTD) or M-Measure was proposed to show the mean distance of pair-wise probability distributions from DNN outputs \cite{hermansky2013mean}. 
Symmetric Kullback–Leibler divergence was selected as a distance metric for distributions $\textbf{p}$ and $\textbf{q}$, which are each posteriors from different time frames. 
\begin{equation}
\label{f:kld}
\mathcal{D}(\mathbf{p}, \mathbf{q}) = \sum_{k=0}^{K} p^{(k)}\log \frac{p^{(k)}}{q^{(k)}} + \sum_{k=0}^{K} q^{(k)}\log \frac{q^{(k)}}{p^{(k)}}
\end{equation}
A high MTD score indicates a greater difference between $\textbf{p}$ and $\textbf{q}$, meaning the model is choosing different output classes at different times.  In noisy conditions or other cases with low model confidence, the distributions at different times should be more similar. 
In MTD, M-Measure often needs to sample frame pairs more than 200 ms apart due to phonetic co-articulation; for shorter time spans, small divergence could be caused by high confidence in the same phoneme. 

In end-to-end framework, we propose Mean Character Distance (MCD), adapted from mean temporal distance. 
Since each probability estimate $\textbf{p}$ in the attention or decoder posterior corresponds to a character prediction, the distance measure is suitable even for adjacent frames without concern for a co-articulation effect. 
So, we take the mean of distance over all pairs from various windows $\{\triangle l\} =\{1,2,3,4,5\}$. 
\begin{equation}
\label{f:mm}
\mathcal{M}_{score} =\frac{\sum_{\{\triangle l\}}\sum_{l=\triangle l}^{L} \mathcal{D}(\textbf{p}_{l-\triangle l}, \textbf{p}_l)}{ \prod_{\{\triangle l\}}(L-\triangle l)}
\end{equation}
Equal weights are applied to all pairs, instead of assigning higher weights for more distant pairs, as with MTD.
As shown in Fig. \ref{fig:mse}(c)(d), similar to the observations for entropy measures, PM on decoder posteriors is better for predicting CERs than PM with attention probabilities. 
Furthermore, MCD is more linearly correlated with CERs than seen with entropy at attention level.

\subsection{Auto-Encoder}

Mean squared error (MSE) of auto-encoder outputs was proposed in \cite{mallidi2015autoencoder} to measure the mismatch between train and test data as an indicator of DNN performance. 
The auto-encoder is trained to minimize the reconstruction error of the DNN pre-softmax activations from training data. 
The previous study illustrated that if a data vector is sampled from training data distribution, the corresponding reconstruction error should be low, while a high error could be observed in a mismatched condition. 

\begin{figure}[tb]
\begin{minipage}[b]{.48\linewidth}
  \centering
  \centerline{\includegraphics[width=4.0cm]{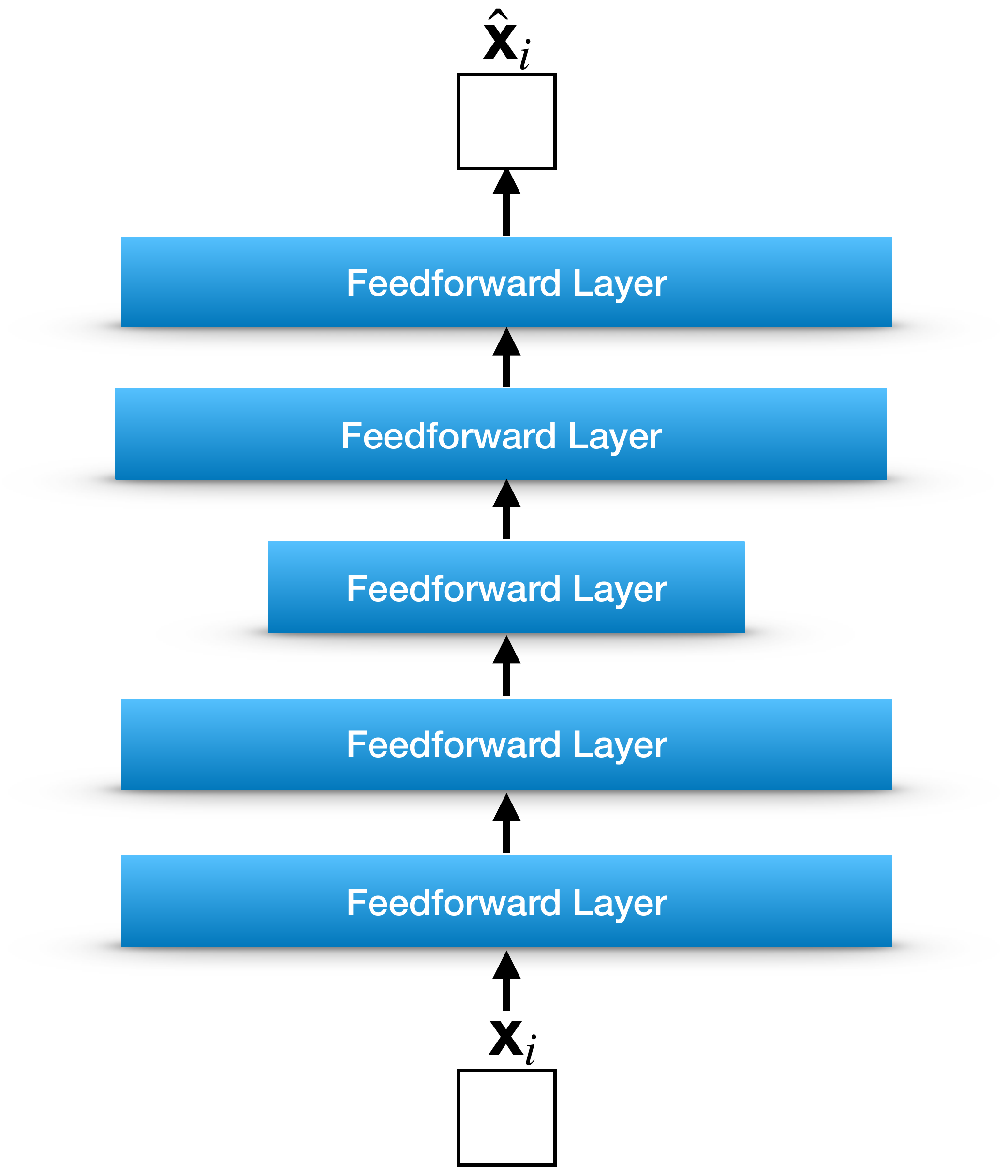}}
%  \vspace{1.5cm}
  \centerline{(a) Auto-encoder}\medskip
\end{minipage}
\hfill
\begin{minipage}[b]{0.48\linewidth}
  \centering
  \centerline{\includegraphics[width=4.0cm]{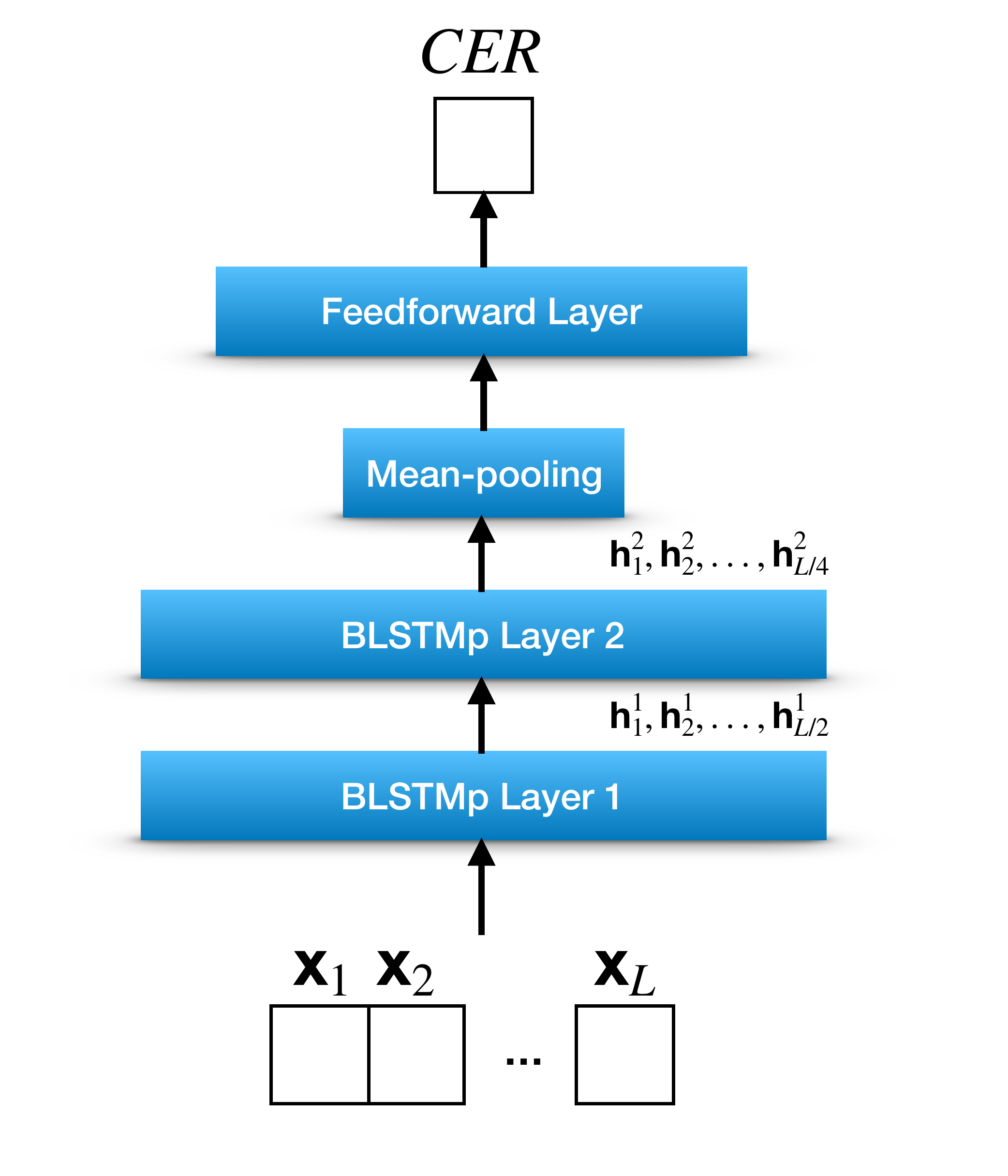}}
%  \vspace{1.5cm}
  \centerline{(b) RNN Predictor}\medskip
\end{minipage}
\caption{Configurations of two PM techniques}
\label{fig:AE_RNN}
\end{figure}

In the end-to-end network, it is natural to apply this technique to 52-dimensional decoder pre-softmax activations. 
The auto-encoder used here is a five-layer 512-unit feed-forward neural network, including a 24-dimensional bottleneck layer in the middle as shown in Fig. \ref{fig:AE_RNN}(a). 
PM score per utterance is derived as average MSE across all frames. 
In the scatter plot Fig. \ref{fig:mse}(e), reconstruction error prediction is consistent with the fitted line for CERs lower than 0.4, while the prediction error diverges for utterances with higher CERs. 
It might be the case that since the auto-encoder only sees the "good" data in training, there is lack of knowledge of how "bad" data looks. 

\subsection{RNN Predictor}
In this work, we propose an RNN-based regression model which directly maps input features of variable length into error rates in the range of CER $[0, +\infty]$.
The model is depicted in Fig. \ref{fig:AE_RNN}(b). 
Two BLSTM layers of 320 units are employed to handle temporal dependencies of inputs. 
Each layer subsamples every other output frame. 
A mean-pooling layer is then used on top of BLSTM outputs to formulate one summary vector per utterance, which is fed into a linear layer of 300 units and an output layer with one Rectified Linear unit (ReLu). 
The model is optimized with MSE loss between predictions and truth CERs. 
Intuitively, the PM score is derived from the model output.
The scatter plot in Fig. \ref{fig:mse}(f) shows that the predictions are well-aligned with the fitted line from linear regression. 
Since it is a direct estimation of CER, the ideal fit should be $CER=PM$. The  derived model using \textit{Dev} is $CER=0.98*PM+0.06$, which is slightly offset and tilted from the ideal case.

\subsection{Overall Results}
Table \ref{tab:mse} summarizes MSEs of linear regression models trained on various PM techniques across different \textit{Test} sets. 
It is worth noting that decoder features work more effectively than attention probabilities in all cases for predicting CERs.
Entropy gives the lowest MSEs for WSJ and Aurora4, domains which have been seen in ASR training. 
The RNN predictor achieves best performance in CHiME4-Sim (not surprising, since this domain was seen in RNN predictor training) as well as real recordings from CHiME4-Real and Dirha-Real, domains not seen in any training stage at all.
MCD measure performs the best at the two unseen simulated domains with reverberant conditions. 
Overall, entropy, MCD, and RNN prediction all provide reasonably good CER predictions, with average prediction errors (square root of MSE) of 10.1\%, 8.8\% and 10.1\%, respectively, where MCD outperforms the rest of PM measures across all test set. 

\begin{table}[tb]
  \caption{Mean Square Error ($\times10^{-2}$) of linear regression trained on performance monitoring techniques. All results are reported on test sets.}
  \label{tab:mse}
  \centering
  \resizebox{0.47\textwidth}{!}{\begin{tabular}{lcccccc}
  \toprule
  \toprule
  & \multicolumn{2}{c}{Entropy} & \multicolumn{2}{c}{MCD}&  &  \\
  Dataset/Domain & Dec& Att& Dec &Att& AE& RNN\\
  \midrule
  {\it Seen (ASR, PM)} \\
  WSJ &\textbf{0.17}&	0.88&	0.25&	0.77&	0.82&	0.29\\
  Aurora4 &\textbf{0.44}	&1.43&	0.47&	1.14&	1.12&	0.74\\
  \midrule
  {\it Seen (PM only)} \\
  CHiME4-Sim &1.13&	5.11& 1.07&	1.87&	2.68&	\textbf{1.01}\\
  \midrule
  {\it Unseen} \\
  CHiME4-Real &1.50&	5.77&	1.24&	2.02&	3.62&	\textbf{1.05}\\
  Reverb-Sim&0.94&	3.42&\textbf{0.78}&	2.20&	1.88&	1.50 \\
  Dirha-Sim &2.23&	6.75&	\textbf{1.16}&	2.05&	6.07&	1.43\\
  Dirha-Real &2.77&	11.80&	1.26&	2.39&	6.97&	\textbf{1.25}\\
  \midrule
  All Together&	 1.02&	3.83&\textbf{0.79}&	1.67&2.49&1.02 \\
  \bottomrule 
  \bottomrule
  \end{tabular}}
\end{table}

\section{Conclusions}
In this work, we investigated four different performance monitoring techniques on attention and decoder features from an end-to-end ASR model. 
Our results show that PM measures on decoder features are more effective for predicting true error rates than PM measures on attention probabilities. 
Entropy and MCD are very simple, effective measures where MCD shows the overall best performance. 
And while auto-encoder methods might be suitable to handle mismatch condition within a certain level of data corruption, an RNN-based regression model shows potential in the direction of performance estimation using deep neural network.  Overall, these results show great promise for performance prediction of end-to-end ASR models.
\bibliographystyle{IEEEtran}

\bibliography{mybib}

\end{document}